\documentclass[10pt, conference, compsocconf]{IEEEtran}

\usepackage{hyperref}
\usepackage{url}
\usepackage{graphicx}
\usepackage{multirow}
\usepackage{subfig}
\usepackage{pbox}
\usepackage{float}

\IEEEoverridecommandlockouts

\begin{document}
\renewcommand{\thefootnote}{\alph{footnote}}

\title{The Royal Birth of 2013:\\Analysing and Visualising Public Sentiment in the UK Using Twitter}

\author{\IEEEauthorblockN{Vu Dung Nguyen, Blesson Varghese\textsuperscript{a} and Adam Barker\textsuperscript{a}}
\IEEEauthorblockA{Big Data Laboratory (\url{http://bigdata.cs.st-andrews.ac.uk})\\
School of Computer Science, University of St Andrews\\
St Andrews, Fife, UK KY16 9SX\\
Email: \{vdn, varghese, adam.barker\}@st-andrews.ac.uk}
}

\maketitle

\begin{abstract}
Analysis of information retrieved from microblogging services such as Twitter can provide valuable insight into public sentiment in a geographic region. This insight can be enriched by visualising information in its geographic context. Two underlying approaches for sentiment analysis are dictionary-based and machine learning. The former is popular for public sentiment analysis, and the latter has found limited use for aggregating public sentiment from Twitter data. The research presented in this paper aims to extend the machine learning approach for aggregating public sentiment. To this end, a framework for analysing and visualising public sentiment from a Twitter corpus is developed. A dictionary-based approach and a machine learning approach are implemented within the framework and compared using one UK case study, namely the royal birth of 2013. The case study validates the feasibility of the framework for analysis and rapid visualisation. One observation is that there is good correlation between the results produced by the popular dictionary-based approach and the machine learning approach when large volumes of tweets are analysed. However, for rapid analysis to be possible faster methods need to be developed using big data techniques and parallel methods. 
\end{abstract}


\IEEEpeerreviewmaketitle
\footnotetext[1]{Corresponding authors}
\renewcommand{\thefootnote}{\arabic{footnote}}

\section{Introduction}
\label{introduction}

Microblogging services such as Twitter have become an important platform for facilitating social interactions in modern society. As demonstrated by recent events such as the Arab Spring and the Occupy Wall Street movements, these platforms can be used to convey powerful ideas and allow the general population to follow such events in real-time. The information posted on these platforms is a rich resource for obtaining insights into the sentiment of the general public. The retrieval and analysis of such information is often referred to as sentiment analysis or opinion mining. 

Traditional methods for understanding public sentiment are questionnaires, surveys and polls which are extremely limited in a number of ways. Firstly, they attract limited participation, and therefore, the sample is not a sufficient representation of the public. Secondly, they are costly to deploy and cannot be used on-the-fly without well laid out logistical plans. Thirdly, they cannot gather the sentiment as an event is unfolding. For example, using traditional methods the sentiment of the people participating in the Occupy Wall Street movement could perhaps be gathered only after the event had finished. 

Currently, Twitter with more than half a billion users is being used as a source for retrieving information. Twitter provides free information through an interface in the form of a stream. Analysis of this information has led to a variety of research. Examples include prediction of elections \cite{predict-1}, stock market \cite{predict-2}, and movie sales \cite{predict-3}, notification of events such as earthquakes \cite{notify-1}, analysis of natural disasters \cite{analysis-1} and public health information \cite{analysis-2}, estimation of public sentiment during elections \cite{pubopinion-1} and recession \cite{pubopinion-2}. This research along with \cite{exemplar-1} are exemplars of how correlated the information retrieved from Twitter and the actual events are. Hence, moving forward a question that arises is - `Why not visualise the information in its geographic context in real-time?'. The research reported in this paper is motivated towards analysing public sentiment related to an event affecting a geographic region in real-time and rapidly visualising it. 

The most common approach employed for analysing public sentiment is dictionary-based \cite{predict-1, predict-2, predict-3} which is simple to implement. Public sentiment, for example, happy, sad or depressed, is understood by comparing tweets against lexicons from dictionaries. A second possible approach that can be employed is machine learning. This approach is not readily available for understanding public (or aggregate) sentiment \cite{machlearn-0a}. However, it is used in understanding the sentiment of individual tweets with high accuracy \cite{machlearn-0b, machlearn-0c}. The research in this paper explores how the machine learning approach can be extended for public sentiment analysis. The notable difference between the two approaches is that the dictionary-based approach classifies individual words in tweets while the machine learning approach classifies an entire tweet. The machine learning approach is quantitatively compared to the dictionary-based approach in this paper. 

The contributions of the research presented in this paper are: (i) the development of a framework for analysing and visualising public sentiment from a Twitter corpus, (ii) the implementation and comparison of two approaches within the framework for analysing public sentiment, (iii) the investigation of visualisation techniques for public sentiment at multiple geographic levels, and (iv) the analysis and visualisation of a Twitter corpus during the birth of Prince George of Cambridge in 2013 as a case study. 

The remainder of this paper is organised as follows. Section \ref{framework} presents a framework for using Twitter to understand public sentiment. Section \ref{casestudies} employs the framework for understanding public sentiment in the UK at the time of the royal birth of 2013. Section \ref{conclusions} concludes this paper by considering future work. 

\section{Framework}
\label{framework}

The framework for analysing and visualising public sentiment presented in this paper can be used to understand the shift of public sentiment seen in tweets and graphically display the sentiment across hours or days or weeks. A score that broadly captures public sentiment is estimated based on two indicators. The first indicator is a positive score to rate how positive the sentiment in a geographic region is. The second indicator is a negative score to rate negative public sentiment in an area. The score can also be normalised with lower and upper bounds as zero and one respectively. The score can be visualised in two geographic levels, namely country and county using a number of visualisation techniques.

\begin{figure}
	\centering
	\includegraphics[width = 0.45\textwidth]{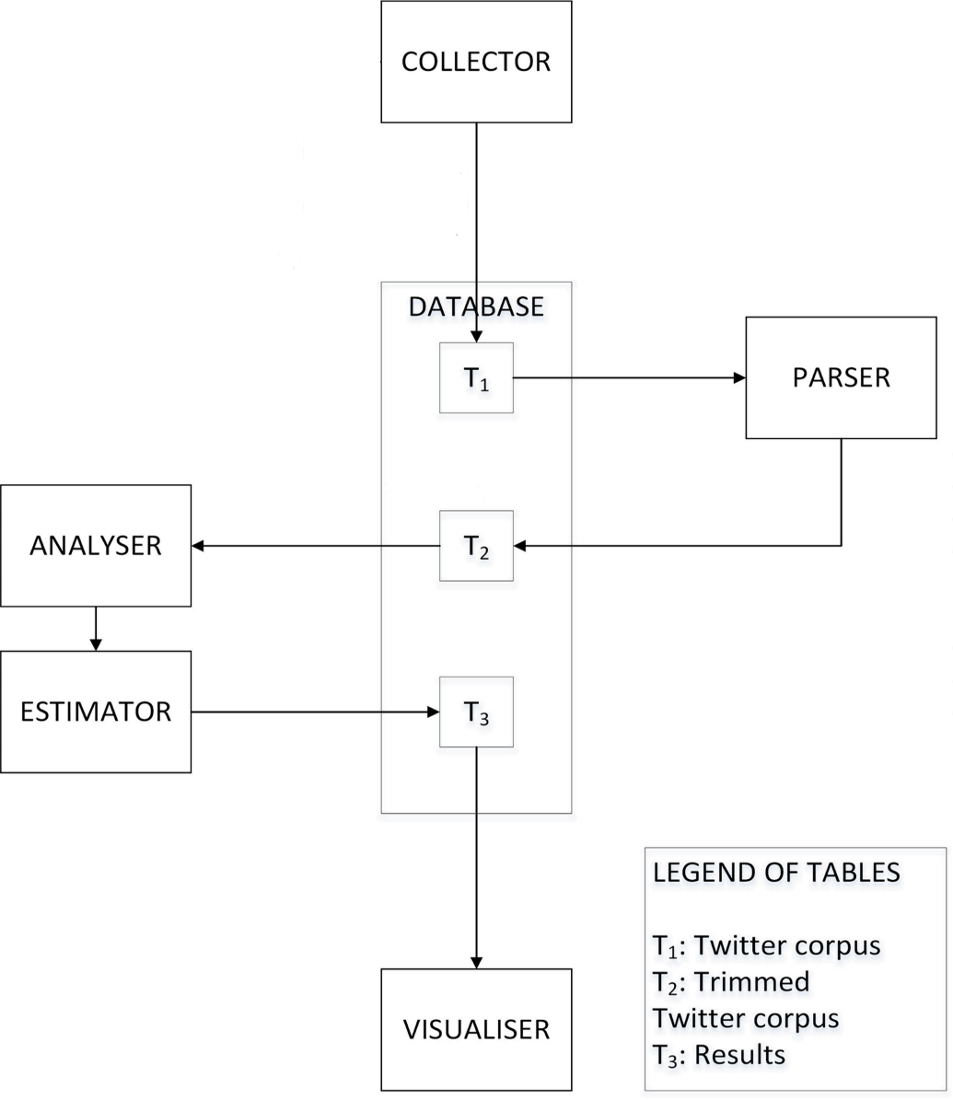}
	\caption{Framework for analysing and visualising public sentiment}
	\label{figure1}
\end{figure}

The framework as shown in Figure \ref{figure1} consists of six modules, namely the Collector, the Parser, the Database, the Analyser, the Estimator and the Visualiser. The Collector module gathers the Twitter corpus. The Parser ensures that the obtained corpus is in a format that can be used by the subsequent modules in the framework. The Database module is a collection of tables containing Twitter data for time periods ranging from minutes to hours to days. The Analyser module mines through the tweets to analyse sentiment. The Estimator module estimates the scores indicating public sentiment. The visualisation of the scores is facilitated through the Visualiser. The flow of data within the framework is also considered in Figure \ref{figure1}. 

\subsection{Collector}
The Collector module is responsible for gathering the Twitter corpus from the Web. The corpus is collected in the JSON format, in real-time, through the Twitter Streaming API\footnote{\url{https://dev.twitter.com/docs/streaming-apis}}. This API not only provides features to select the geographic region of the tweets' origin but also provides options to select parameters such as keywords and language. 

\subsection{Parser}
The Parser module is essential to trim the corpus offline. The collection and trimming operations are performed in two different stages since the Twitter Streaming API provides tweets at a fast rate. Parsing the corpus in real-time may cause the tweets that are streamed to be lost if the Parser cannot keep up with the data flow of the Streaming API. The output from the Parser makes the corpus readable for the subsequent modules in the framework.

\subsection{Database}
The Database module consists of three tables shown as $T_{1}$, $T_{2}$ and $T_{3}$ in Figure 1. $T_{1}$ is the tweet corpus gathered by the Collector. $T_{1}$ is then parsed to produce $T_{2}$, a trimmed readable table. The Analyser retrieves data from $T_{2}$ for analysis and the Estimator writes $T_{3}$ containing the public sentiment scores and associated geographic and time information. 

\subsection{Analyser}
This module performs sentiment analysis to extract the sentiment of the tweets. Two approaches are explored in this paper for performing sentiment analysis, namely the dictionary-based and machine learning approaches. The aim of both the approaches is to estimate a score that captures the degree of `positive' or `negative' public sentiment of a geographic region in a time frame by evaluating a collection of tweets or individual tweets. The dictionary-based approach considers the entire collection of tweets for a given time period to aggregate the public sentiment across the collection. However, in the machine learning approach each tweet in the collection is assigned a sentiment score and then the public sentiment is aggregated from individual scores. The public sentiment score generated by both the approaches is independent of the number of tweets. 

\subsubsection{Approach 1 - Dictionary-based}
\begin{figure}
	\centering
	\subfloat[Dictionary-based]{\label{figure2a}\centering \includegraphics[width=0.27\textwidth]{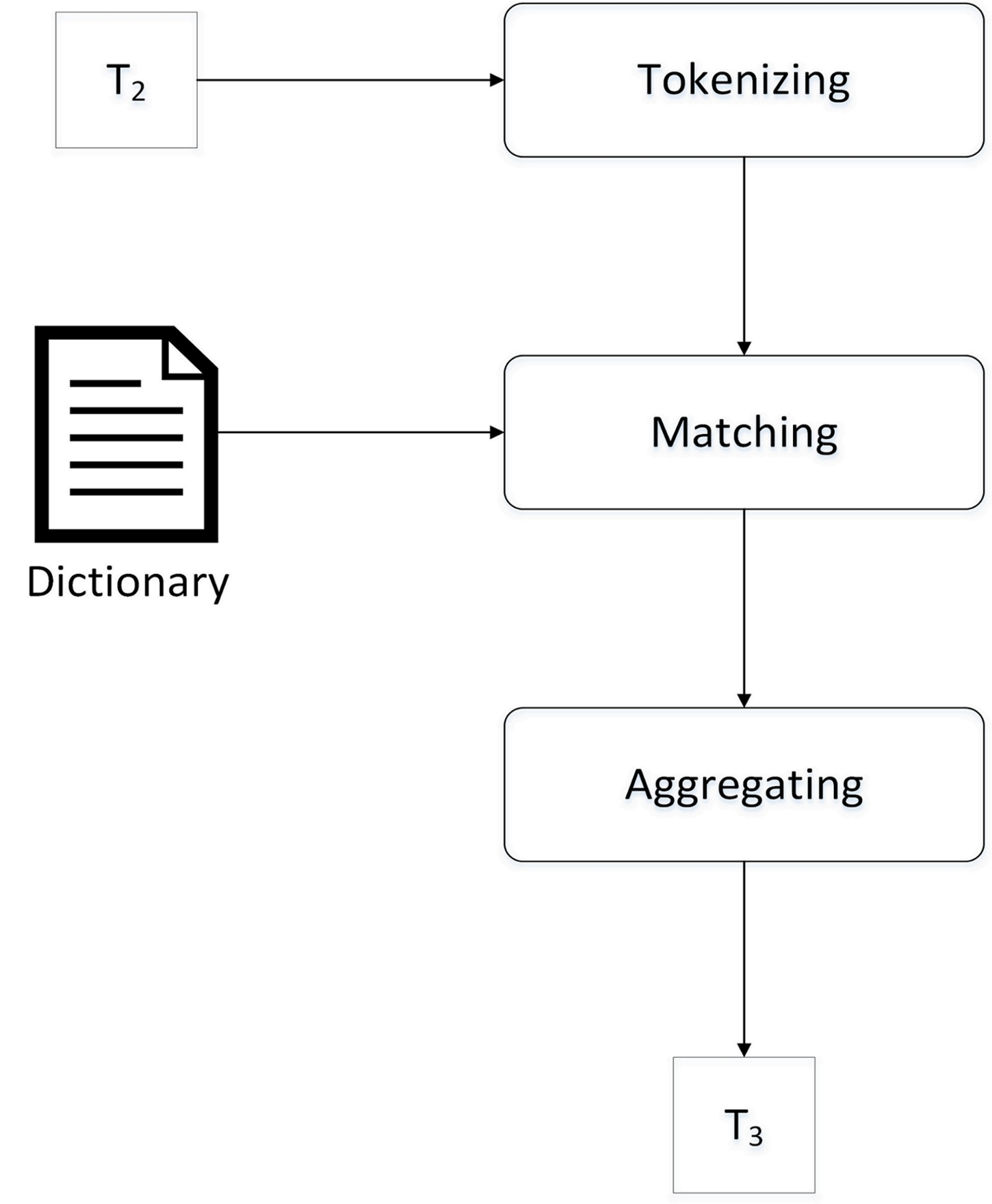}} \\
	\subfloat[Machine learning]{\label{figure2b}\centering \includegraphics[width=0.35\textwidth]{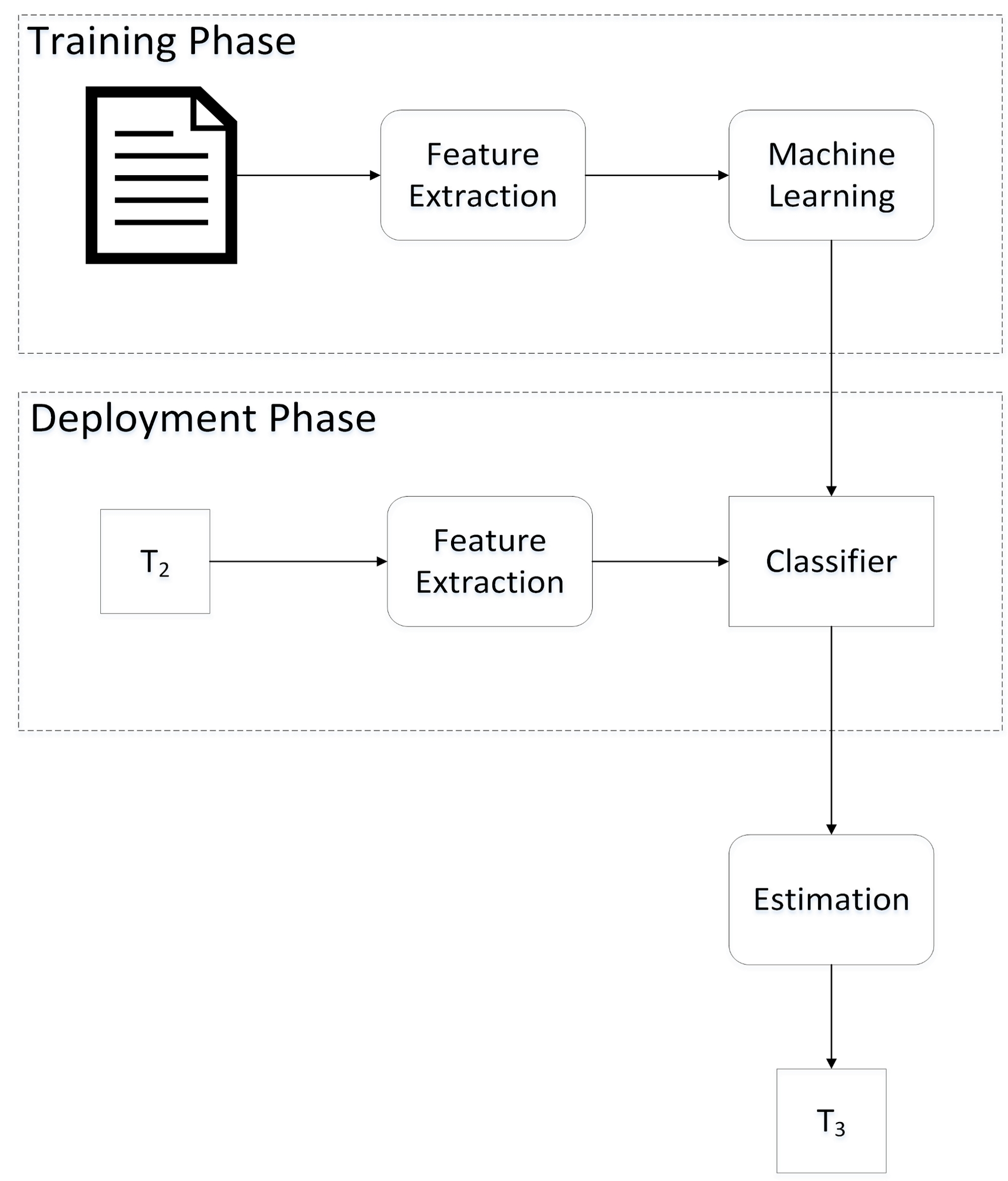}}
\caption{Sentiment analysis appraoches}
\label{Figure2}
\end{figure}

Figure \ref{figure2a} shows the dictionary-based approach. The input is a data set selected for a time period from a specified geographic region (for example, country or county). The tweets of the selected data set are tokenised using a lexical analyser. For this the Stanford tokeniser \cite{dic-1a, dic-1b} which incorporates the Penn Treebank 3 (PTB) tokenisation algorithm \cite{dic-2} is employed. The tokens are then matched against a dictionary; the Emotional Lookup Table provided by SentiStrength \cite{dic-3a, dic-3b} is used as the dictionary. While matching, the number of positive sentiment and negative sentiment words in the entire set of tokens are counted. Then the public sentiment is aggregated by calculating the ratio of the positive sentiment to negative sentiment words. 

\subsubsection{Approach 2 - Machine Learning}
Figure \ref{figure2b} shows the machine learning approach. In contrast to the dictionary-based approach in which `prior linguistic knowledge' in the form of dictionaries were used, the machine learning approach implemented in this paper considers a supervised training technique. The machine learning approach is presented in three phases - firstly, the training phase, secondly, the testing phase, and finally, the deployment phase. 

In the training phase, the training data was collected using the approach presented in \cite{machlearn-1} which relies on the Distant Supervisor technique \cite{machlearn-2}. The training data set contains 23,000 tweets which are labelled as positive or negative. This approach is in contrast to the manual approach reported in \cite{machlearn-3, machlearn-4} which requires human intervention for labelling tweets. Unigram features are extracted from the training data set to train the classifier model; the Naive Bayes Classifier model is used. 

After training the model, in the testing phase, the approach is tested using the data set available from \cite{machlearn-5}. The test results indicate over 70\% accuracy in labelling tweets and a similar finding is reported in \cite{machlearn-1, machlearn-6}.

In the deployment phase, the tweets for a geographic region are selected from the table containing parsed tweets, $T_{2}$. These tweets are labelled using the Classifier obtained from the training phase. The number of positive sentiment and negative sentiment tweets in the entire collection of tweets is counted, and public sentiment is then aggregated by calculating the ratio of the positive sentiment to negative sentiment tweets. 

\subsection{Estimator}
The Estimator module computes a score that captures public sentiment. The estimation technique employed in the dictionary-based approach is subtly different from the machine learning approach and is considered in this section. 

\subsubsection{Estimation in the dictionary-based approach}
Consider a geographic region defined by $g = 1 \mbox{ and } 2$, where $g = 1$ for a country and $g = 2$ for a county and time frame $t$. The public sentiment score is defined as:

\begin{equation}
PSS_{(g,t)} = \frac{count_{(g,t)}\big(\mbox{positive words}\big)}{count_{(g,t)}\big(\mbox{negative words}\big)} \nonumber
\end{equation}

\begin{figure*}
	\centering
	\includegraphics[width = 0.7\textwidth]{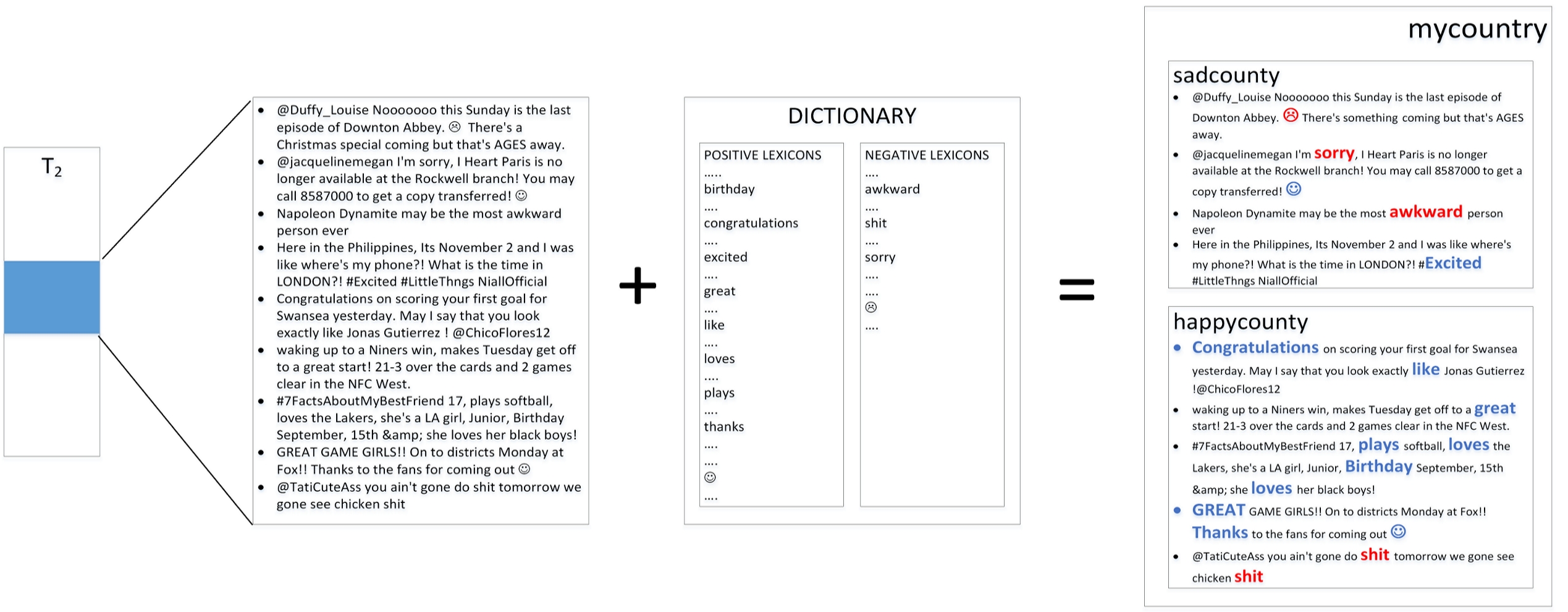}
	\caption{Illustration of an example using dictionary-based approach}
	\label{figure3}
\end{figure*}

The example illustrated in Figure \ref{figure3} for a geographic region has one country, $mycountry$ with two counties $happycounty$ and $sadcounty$. The tweets for the region are selected from the table containing parsed tweets, $T_{2}$, for a time frame denoted as $t$, starting at $t_{start}$ and ending at $t_{end}$. The selected data during the time frame is represented in the figure as a collection of nine tweets, five from $happycounty$ and four from $sadcounty$. The tweets are then matched against a dictionary which results in the recognition of positive and negative words. In the figure, the positive words are represented in blue and the negative words in red. The number of positive words in the tweets is twelve (ten from $happycounty$ and two from $sadcounty$) and the number of negative words is five (two from $happycounty$ and three from $sadcounty$). Therefore, the public sentiment score for time $t$ at country level for $mycountry$ is 2.4, and the public sentiment score at the county level for $happycounty$ is 5.0 and $sadcounty$ is 0.66. The scores for the counties can be normalised between 0 and 1, and so the normalised public sentiment score is 1.0 for $happycounty$ and is 0.132 for $sadcounty$. Geographic distinctions (counties) can highlight the finer level of detail which can be lost when aggregated to higher geographic level (country). 

\subsubsection{Estimation in the Machine Learning approach}
Consider a geographic region defined by $g = 1 \mbox{ and } 2$, where $g = 1$ for a country and $g = 2$ for a county and time frame $t$. The public sentiment score is defined as:

\begin{equation}
PSS_{(g,t)} = \frac{count_{(g,t)}\big(\mbox{positive tweets}\big)}{count_{(g,t)}\big(\mbox{negative tweets}\big)} \nonumber
\end{equation}

\begin{figure}
	\centering
	\includegraphics[width = 0.5\textwidth]{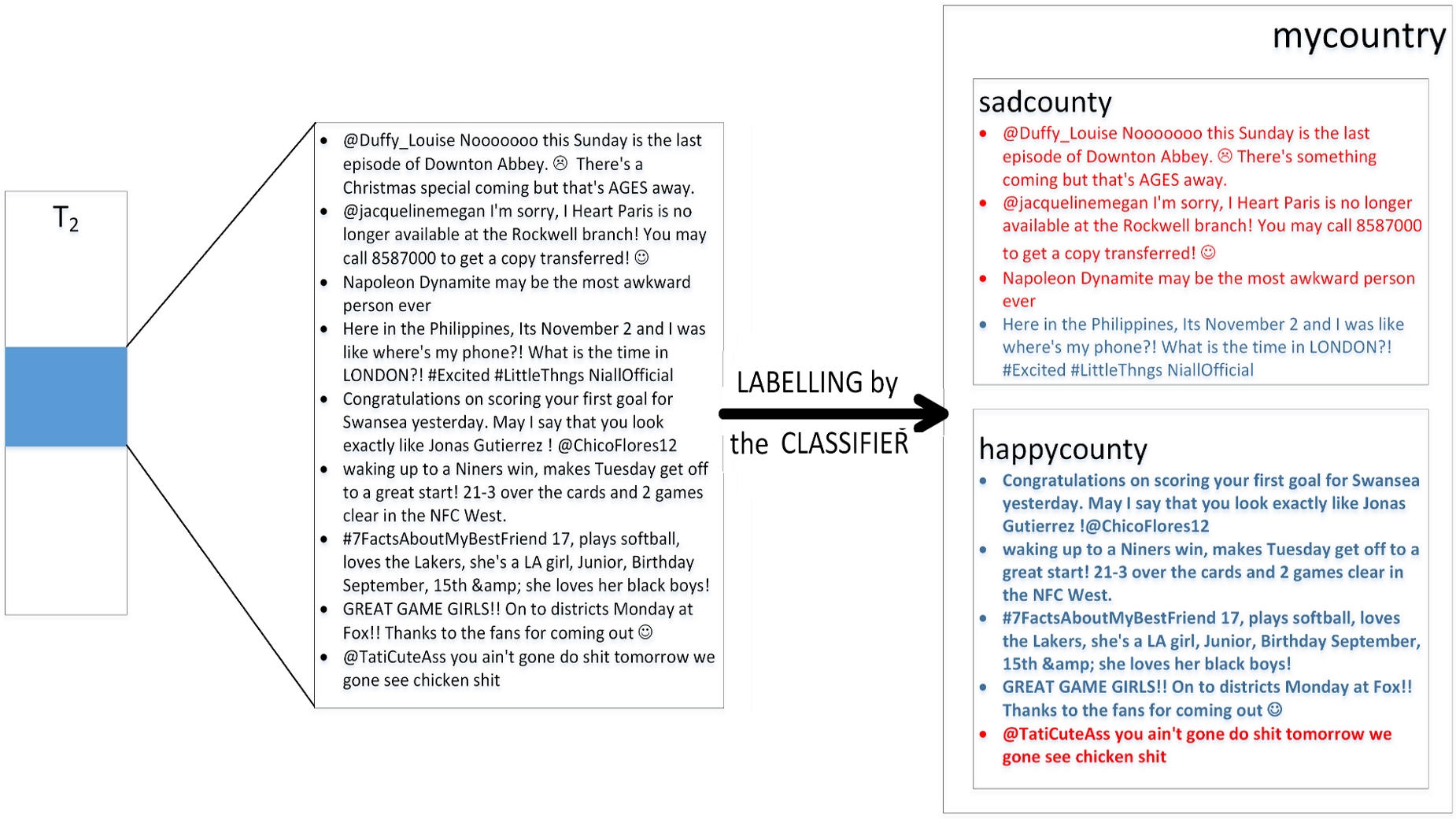}
	\caption{Illustration of an example using machine learning approach}
	\label{figure4}
\end{figure}

The example illustrated in Figure \ref{figure4} for a geographic region has one country, $mycountry$ with two counties $happycounty$ and $sadcounty$. The tweets for the region are selected from the table containing parsed tweets, $T_{2}$, for a time frame denoted as $t$, starting at $t_{start}$ and ending at $t_{end}$. The selected data during the time frame is represented in the figure as a collection of nine tweets, five from $happycounty$ and four from $sadcounty$. The classifier labels the tweets as positive sentiment and negative sentiment. In the figure, the positive tweets are represented in blue and the negative tweets in red. The number of positive tweets is five (four from $happycounty$ and one from $sadcounty$) and the number of negative tweets is four (one from $happycounty$ and three from $sadcounty$). Therefore, the public sentiment score for time $t$ at country level for $mycountry$ is 1.25, and the public sentiment scores at the county levels for $happycounty$ and $sadcounty$ are 4 and 0.33 respectively. The normalised public sentiment score between 0 and 1 for the counties are 1.0 for $happycounty$ and 0.0825 for $sadcounty$.

The $PSS$ score from both approaches are normalised to $NPSS$ to be able to compare the public sentiment trend estimated by the approaches. 

\subsection{Visualiser}

The Visualiser module facilitates the graphical display of public sentiment using three visualisation techniques. The first technique is choropleth visualisation of public sentiment on a geo-browser. In the research reported in this paper, Google Earth\footnote{\url{http://earth.google.co.uk/‎}} is employed as the geo-browser. The Thematic Mapping Engine (TME) \cite{vis-1} is used for generating .kml files \cite{vis-2} in which public sentiment data overlays geographic data. Choropleth is useful for presenting public sentiment as a gradient of colours, and in this framework the public sentiment of a country is presented using choropleth. For example, the public sentiment of England, Scotland, Wales and N. Ireland is represented by overlaying colours indicative of public sentiment in each country over the geographic region on Google Earth. Public sentiment of counties are not best represented using choropleths since it would be visually difficult to distinguish between colours overlaid on small geographic regions. While multiple dimensions of data can be represented using distinct gradient scales it may be visually challenging to distinguish between the scales. 

The second technique using tile-maps is independent of a geo-browser. A geographic region is represented as a tile and the public sentiment of the region can be visually distinguished not only based on the colour of the tile but also on its size. Google Charts API \footnote{\url{https://developers.google.com/chart/}} is used for obtaining tile-maps in the framework. For example, the public sentiments of all the counties in the UK are represented using tiles.

The third technique using line graph visualisation is again independent of a geo-browser. This technique is useful to understand the relative performance of the two sentiment analysis approaches over the dimension of time. For example, the public sentiment in England in the hour following the announcement of Prince George's birth, estimated using the dictionary-based approach and the machine learning approach, can be compared and represented using line graphs.

\section{Case Study: UK Royal Birth, 2013}
\label{casestudies}
The royal birth of Prince George of Cambridge on Monday, 22 July, 2013 at 16:24 BST to the Duke and Duchess of Cambridge is considered in the framework for analysing and visualising public sentiment. The first Twitter announcement on the day of birth that the arrival of the baby was soon expected was made at 07.37 BST. This attracted a lot of attention from Twitter users in the UK and across the world. Nearly 487 million users accessed tweets related to the birth\footnote{\url{http://www.dailymail.co.uk/news/article-2374252/Royal-babys-birth-\\news-sends-Twitter-meltdown-487m-congratulate-Duchess-Cambridge.html}}. This section considers the pipeline of activities to analyse the tweet corpus, followed by visualising the results obtained from the analysis, and finally, summarises the key observations from the case study.

\begin{figure*}
	\centering
	\subfloat[Dictionary-based Approach - 21 July, 2013]{\label{figure5-1a}\centering \includegraphics[width=0.325\textwidth]{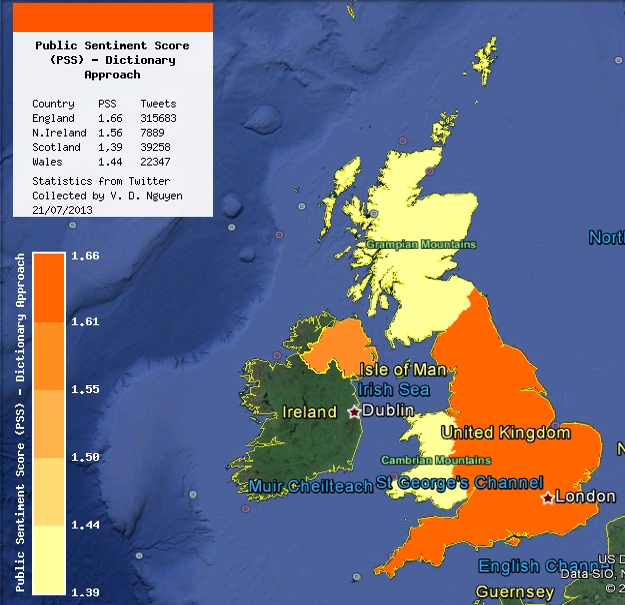}}\hfill
	\subfloat[Dictionary-based Approach - 22 July, 2013]{\label{figure5-1b}\centering \includegraphics[width=0.325\textwidth]{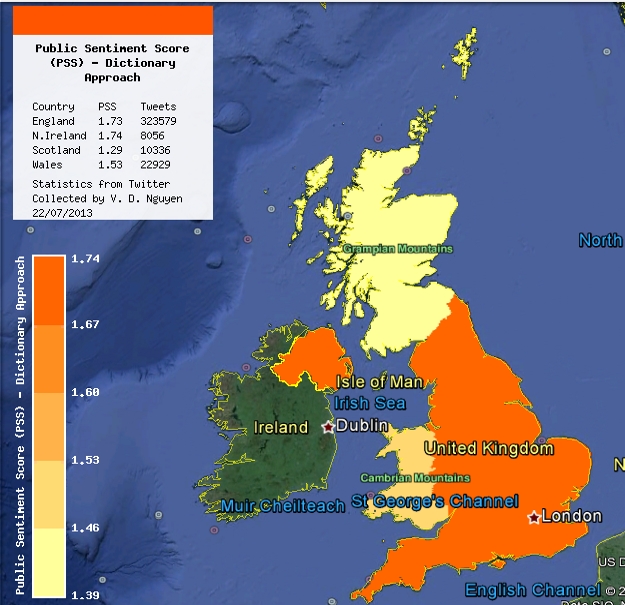}} \hfill
	\subfloat[Dictionary-based Approach - 23 July, 2013]{\label{figure5-1c}\centering \includegraphics[width=0.325\textwidth]{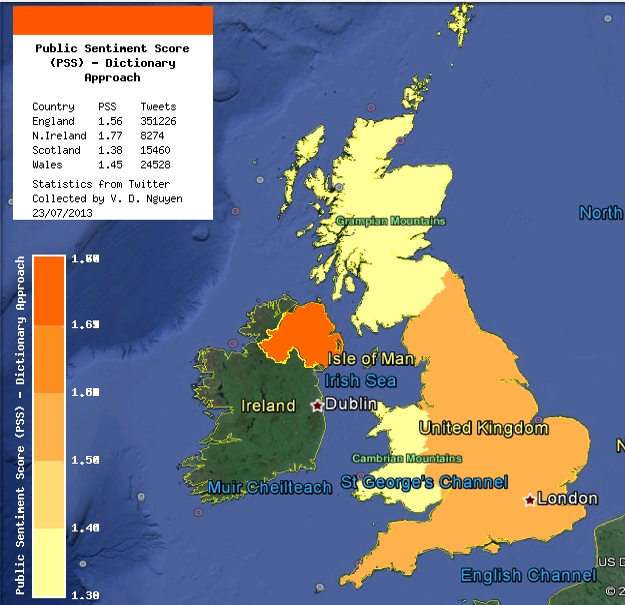}} \\

	\subfloat[Machine Learning Approach - 21 July, 2013]{\label{figure5-2a}\centering \includegraphics[width=0.325\textwidth]{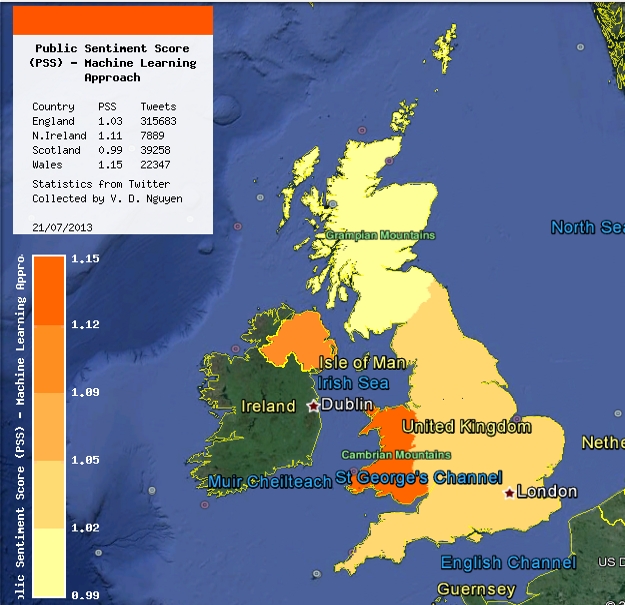}} \hfill
	\subfloat[Machine Learning Approach - 22 July, 2013]{\label{figure5-2b}\centering \includegraphics[width=0.325\textwidth]{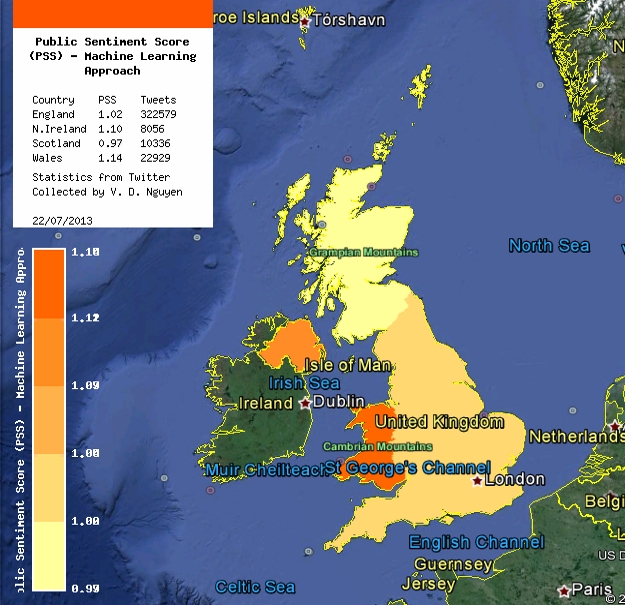}} \hfill
	\subfloat[Machine Learning Approach - 23 July, 2013]{\label{figure5-2c}\centering \includegraphics[width=0.325\textwidth]{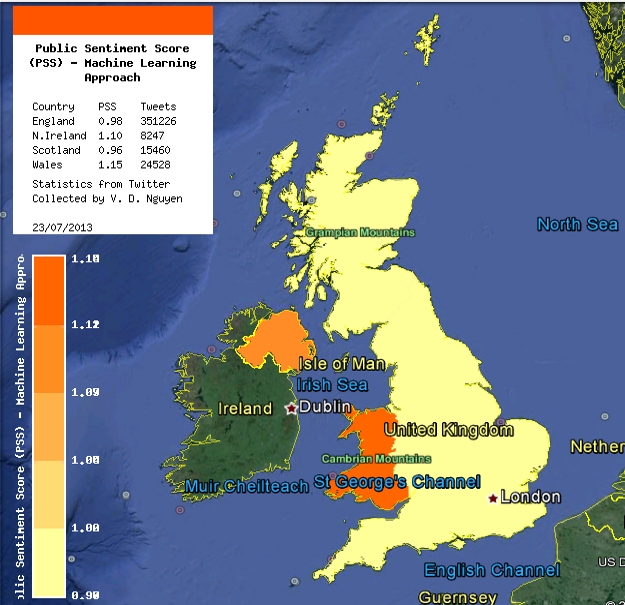}} 
\caption{Public Sentiment Score of England, Wales, Scotland and N. Ireland for case study}
\label{Figure5}
\end{figure*}

\begin{table*}
	\centering
	\begin{tabular}{| l | r | c c r r | c c r r |}
		\hline
		\multirow{2}{*}{\textbf{Country}} & \multirow{2}{*}{\textbf{\pbox{2.2cm}{No. of\\Tweets}}} & \multicolumn{4}{|c|}{\textbf{Dictionary-based Approach}} & \multicolumn{4}{|c|}{\textbf{Machine Learning Approach}}\\
		\cline{3-10}
			&	& \textbf{NPSS}	& \textbf{PSS}	& \textbf{\pbox{2.5cm}{No. of\\Positive Words}}	&\textbf{\pbox{2.5cm}{No. of\\Negative Words}}	& \textbf{NPSS} & \textbf{PSS}	& \textbf{\pbox{2.5cm}{No. of\\Positive Tweets}}	&\textbf{\pbox{2.5cm}{No. of\\Negative Tweets}} \\
		\hline
		\hline		
		\multicolumn{10}{|c|}{21 July 2013}\\
		\hline
		England				&315,658 & 1.0000 & 1.6620 & 166,607 	& 100,244	& 0.8925 & 1.0270 & 159,928 	& 155,730 	\\ 
		Scotland			&39,233	 & 0.8351 & 1.3880 & 20,384 	& 14,685	& 0.8630 & 0.9930 & 19,548 	& 19,685	\\ 
		Wales				&22,322	 & 0.8688 & 1.4439 & 11,379 	& 7,881		& 1.0000 & 1.1507 & 11,943 	& 10,379	\\ 
		N. Ireland			&7,864	 & 0.9401 & 1.5625 & 4,389 	& 2,809		& 0.9666 & 1.1123 & 4,141 	& 3,723		\\ 
		\hline
		\multicolumn{10}{|c|}{22 July 2013}\\
		\hline
		England				&322,554 & 1.0000 & 1.7398 & 176,784 	& 102,189	& 0.9648 & 1.0992 & 162,986 	& 159,568 	\\ 
		Scotland			&10,312	 & 0.7980 & 1.3884 & 5,247 	& 3,779		& 0.8502 & 0.9686 & 5,074 	& 5,238		\\ 
		Wales				&22,904	 & 0.8794 & 1.5301 & 12,522 	& 8,184		& 1.0000 & 1.1392 & 12,197 	& 10,707	\\ 
		N. Ireland			& 8,031	 & 0.9943 & 1.7299 & 4,755 	& 2,733		& 0.8966 & 1.0214 & 4,205 	& 3,826		\\ 
		\hline
		\multicolumn{10}{|c|}{23 July 2013}\\
		\hline
		England				&351,201 & 0.8801 & 1.5621 & 188,931 	& 120,948	& 0.8535 & 0.9824 & 174,045 	& 177,156 	\\ 
		Scotland			&13,816	 & 0.7771 & 1.3793 & 7,509 	& 5,444		& 0.8460 & 0.9734 & 6,815 	& 7,001		\\ 
		Wales				&24,233	 & 0.8166 & 1.4493 & 13,039 	& 8,997		& 1.0000 & 1.1510 & 12,967 	& 11,266	\\ 
		N. Ireland			&8,222	 & 1.0000 & 1.7749 & 4,755 	& 2,679		& 0.9581 & 1.1028 & 4,312 	& 3,910		\\ 
		\hline
	\end{tabular}
	\caption{Summary of results from case study}
	\label{table1a}
\end{table*}

\subsection{Analysing the tweets}
The Twitter corpus was being collected for the UK by the Collector module using the Twitter Streaming API from Sunday, July 21 2013, 00:00:01 BST until Tuesday, 23 July, 2013, 23:59:59 BST. Nearly one million tweets were collected from over 150,000 Twitter users. The case study is used to compare the dictionary-based and machine learning approaches. The geographic area taken into account is the UK. 

The Parser module trimmed the corpus, and the fine level of geographic details, namely latitude and longitude, was used to map the tweets onto the county and the country of origin using the Global Administrative Areas (GADM) spatial database\footnote{\url{http://www.gadm.org}} as shapefiles (.shp) \cite{shapefile-1}. The dictionary-based and machine learning approaches were used for sentiment analysis and the aggregation of public sentiment was performed. The results obtained at the country level for July 21, July 22 and July 23 are summarised in Table \ref{table1a}, where PSS is the Public Sentiment Score and NPSS is the normalised PSS. 

\subsection{Visualisation}
Three techniques presented in Section \ref{framework} are considered for visualising the public sentiment in the UK. They are firstly, the choropleth visualisation technique is overlaid on Google Earth for the country level, secondly the tile-map visualisation technique for the county level, and thirdly, the line graph visualisation technique on a hourly basis at the country level. 

\subsubsection{Visualisation on geo-browser}
Figure \ref{Figure5} shows screenshots of PSS using choropleth visualisation on Google Earth for July 21, July 22 and July 23 based on Table \ref{table1a}. The highest volume of tweets was obtained from England, followed by Wales and then Scotland. The smallest number of tweets during the three day period was from N. Ireland. On July 22 and July 23 the dictionary-based approach estimates England to have had the highest PSS compared to the other countries. Surprisingly, on the day after the birth, England dropped to the third place. On the other hand, the machine learning approach places England consistently in third place. The machine learning approach estimates Wales to have the highest PSS on all three days. 

Further, a correlation analysis between the PSS obtained from both the approaches was performed. The results obtained are summarised in Table \ref{table1b}, where the correlation ratio indicates the closeness of the PSS scores estimated by the dictionary-based and machine learning approaches. Given the large volume of tweets analysed for England, there is a large correlation of over 80\% between the results produced by both the approaches. The two approaches produce least correlated results for Wales, and the correlation ratios for Scotland and N. Ireland are not high. This is perhaps because the analysis on larger volumes of tweets can produce higher quality of results. 

\begin{figure*}
	\centering
	\subfloat[Dictionary-based approach]{\label{figure6a}\centering \includegraphics[width=0.495\textwidth]{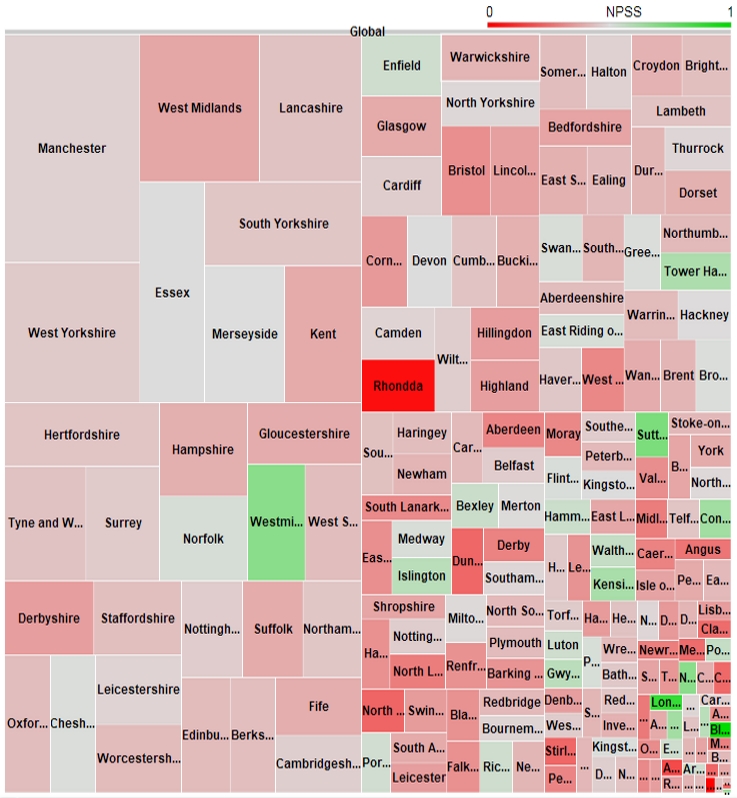}}
	\subfloat[Machine learning approach]{\label{figure6b}\centering \includegraphics[width=0.495\textwidth]{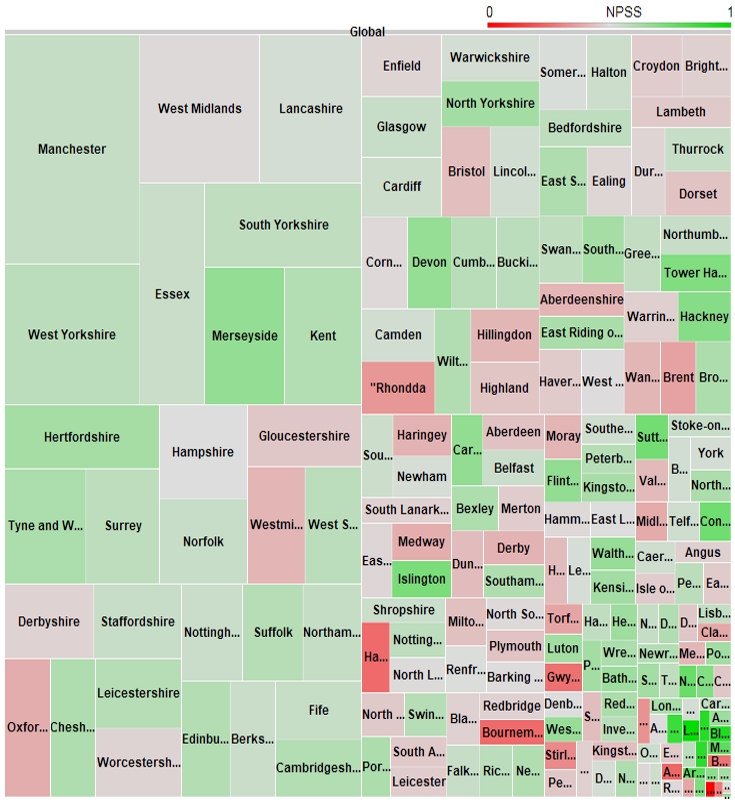}} 
\caption{Tile-map representation of Public Sentiment Score in UK counties}
\label{Figure6}
\end{figure*}

\subsubsection{Visualisation using tile-maps}
Figure \ref{Figure6} shows the tile-map representation of the NPSS corresponding to all UK counties using the dictionary-based approach and machine learning approach. Each tile represents a county, and the size of each tile is relative to the volume of tweets that originated from the county. The colour of the tile is indicative of the normalised PSS varying from shades of red (lowest NPSS score) to green (highest NPSS score). The largest volume of tweets is from Manchester, West Yorkshire, West Midlands, Lancashire, Essex all in England, and the lowest volume is from Strabane, Larne and Moyle in N. Ireland, Rhonndda in Wales, Orkney Islands and Shetland Islands. Using the dictionary-based approach the public sentiment score is highest for the Greater London area that includes London, Sutton, Westiminster, Kensington and Chelsea, Tower Hamlets and Islington, and is the lowest for Shetland Islands, Armagh in N. Ireland and Rhondda in Wales. There is a predominance of the red shade and this is largely because there are relatively few high PSS values. Therefore, when the lower PSS values are normalised using the approach presented in Section \ref{framework} they diminish greatly.

The trends seen in the dictionary-based approach are quite comparable to the trends seen in the machine learning approach. Using the machine learning approach Strabane, Shetland Islands and Rhondda have very high PSS scores which are notable exceptions. This is so because a very small number of tweets are analysed for these counties. Surprisingly, Rhondda falls under the exception though there is a reasonably large volume of tweets. Similar to the dictionary-based approach, Larne has the a low NPSS in the machine learning approach. The regions that had a high NPSS score in the dictionary-based approach are also found to have a high NPSS score using machine learning. 

\subsubsection{Visualisation using line graphs}
\begin{figure}
	\centering
	\subfloat[England	]{\label{figure7a}\centering \includegraphics[width=0.5\textwidth]{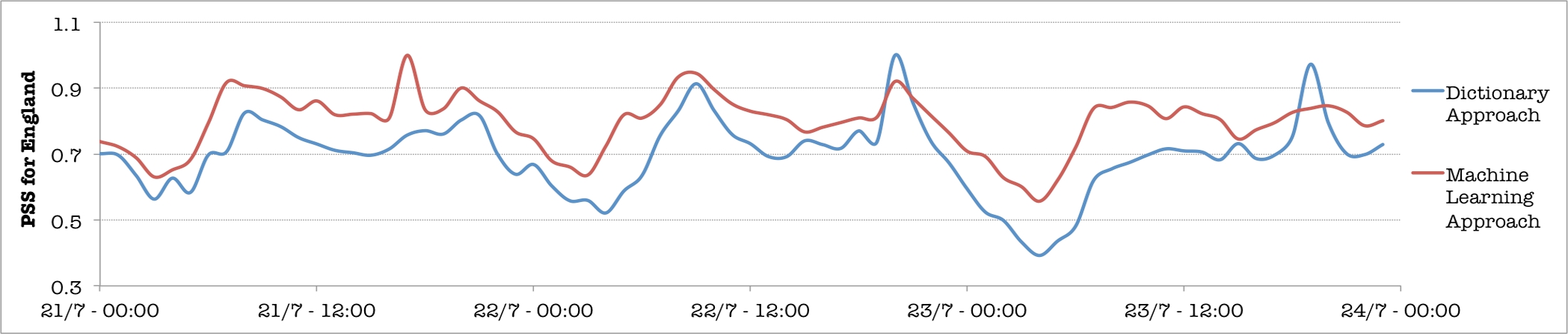}} \\
	\subfloat[Wales		]{\label{figure7b}\centering \includegraphics[width=0.5\textwidth]{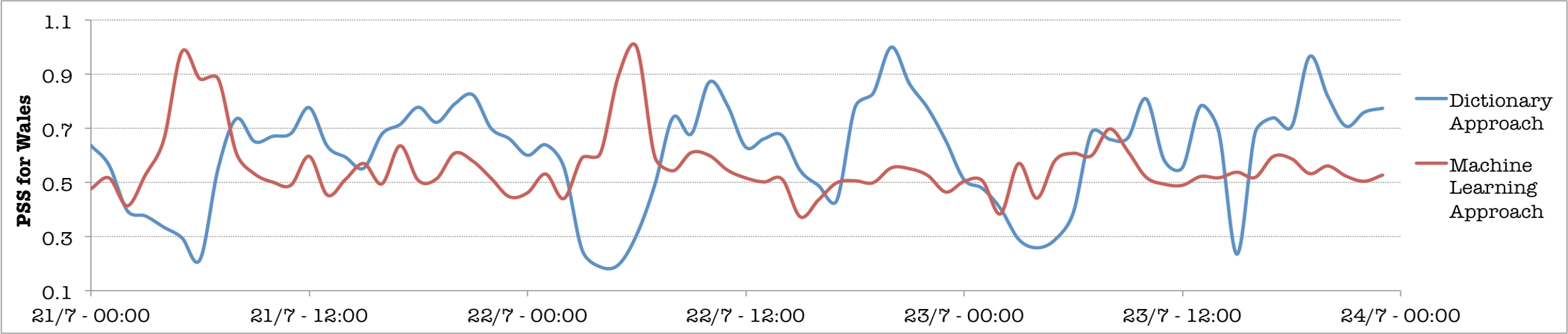}} \\
	\subfloat[N. Ireland	]{\label{figure7c}\centering \includegraphics[width=0.5\textwidth]{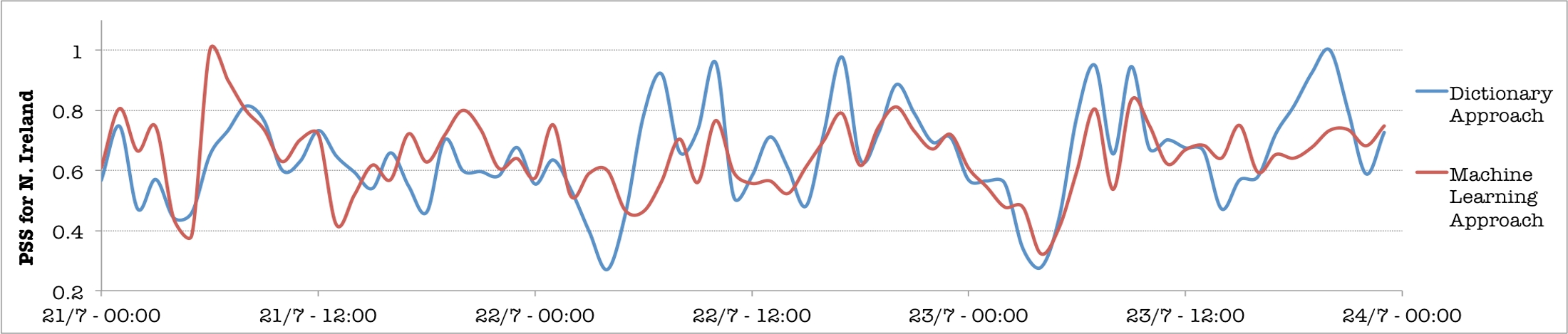}} \\
	\subfloat[Scotland	]{\label{figure7d}\centering \includegraphics[width=0.5\textwidth]{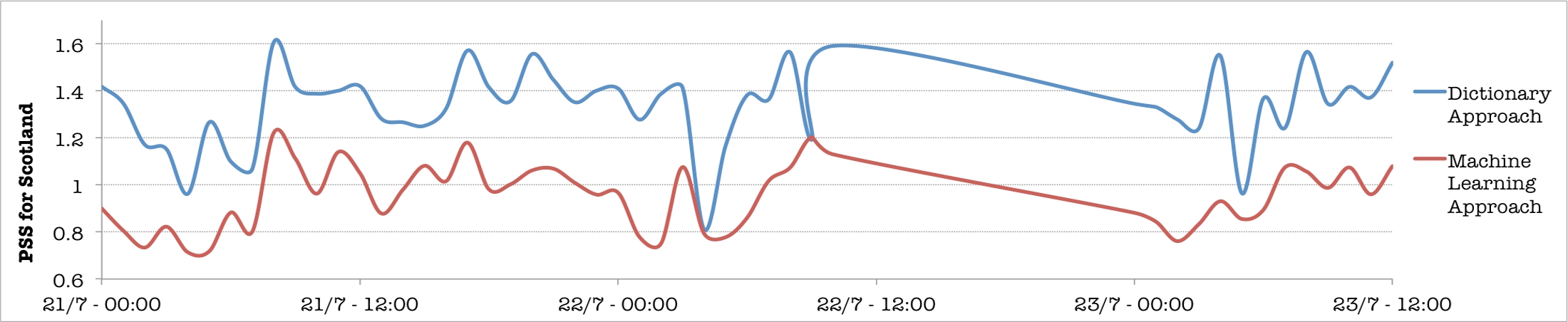}} \\
\caption{Variation of public sentiment in the UK from 21-23 July 2013}
\label{Figure7}
\end{figure}

Figure \ref{Figure7} shows the visualisation of the trend of public sentiment in England, Wales, N. Ireland and Scotland from 21 July 2013 to 23 July 2013. The tweet corpus for Scotland after 10:00 BST was not obtained on 22 July 2013. The number of tweets used to analyse the sentiment for England was nearly one million, for Wales was over 69,000, for N. Ireland was over 24,000, and for Scotland was nearly 65,000. In general, both the dictionary-based and machine learning approaches produce the same trend though several exceptions can be noted; in the case of England, there seems to be fewer exceptions and is likely to be because a large number of tweets are analysed. For Wales the exceptions are seen for two time periods, firstly, between 00:00 and 07:00, and secondly, between 17:00 to 20:00. Though the dictionary-based approach estimates an increasing positive trend in the sentiment score after the birth of the Prince, the machine learning approach fails to capture this. In the case of N. Ireland there is a close similarity in the trend between 22 July 12:00 BST and 23 July 12:00 BST when there was a high volume of tweets regarding the birth. Similarity in the increasing and decreasing trends of PSS across the days are also noted for Scotland. 

\begin{table}
	\centering
	\begin{tabular}{| l | r | c |}
		\hline
		\textbf{Country}	&	\textbf{No. of Tweets}			& 	\textbf{Correlation Ratio}\\
		\hline
		\hline
		England			&	989,413					& 	0.8192\\
		Scotland		&	64,980					&	0.6110\\
		Wales			& 	69,459					&	0.3146\\
		N. Ireland		&	24,117					&	0.5485\\
		\hline
	\end{tabular}
	\caption{Correlation ratio between the dictionary-based and the machine learning approaches}
	\label{table1b}
\end{table}

\subsection{Discussion}
In the case of England, during the announcement of the birth on July 22 and for a few hours later the PSS has a steady trend at an average of 0.7. This indicates that the tweets posted during this time have nearly 30\% more negative sentiments than positive sentiments. However, after 20:00 BST on July 22 there is a quick spike in the PSS lasting a couple of hours which is again noted on July 21 and July 23. This is perhaps due to the increase in the volume of tweets posted during these hours. Interestingly, for Wales and N. Ireland an increasing trend with higher PSS scores are noted. For example, using the dictionary-based approach in Wales a steady rise of the PSS from less than 0.5 to over 1.0 is noted during and after the birth. Since this trend is not observed the previous day or the day after the birth it can be inferred that the people of Wales were more positive during the time of the birth than the people in England. A progressively steady decrease is noted in the public sentiment of Scotland, though the PSS during and after the time of the birth is higher than that of England. 

In summary, inspite of the fact that there is strong correlation between the two approaches for England, the dictionary approach places England in the first place for July 21 and July 22 and then in the third place for July 23, and the machine learning approach places England in the third place in the UK from 21-23 July for positive public sentiment. Therefore, `Does England react quickly to events unlike other member countries?' This is a pointer to further investigation and is beyond the scope of this paper. 

To conclude, the case study indicates that the public sentiment scores estimated by the machine learning approach is highly correlated to the dictionary-based approach when large volumes of tweets are analysed for a time period. Nonetheless, several exceptions are noted and will require a closer investigation. While the current implementation of the machine learning approach is slow it is possible to be employed for offline estimation, particularly when an analysis of a past event is being performed. Case studies to validate the use of the framework for analysing past events will be reported elsewhere.  

\section{Conclusions}
\label{conclusions}

This paper presented a framework for the analysis and visualisation of public sentiment. The framework comprises modules to collect, parse, analyse, estimate and visualise the estimated public sentiment. A Public Sentiment Score (PSS) and a normalised PSS based on positive and negative indices that broadly capture public sentiment of geographic regions was used in this research. The scores were graphically visualised on a geo-browser, as tile-maps and as time graphs. The two underlying approaches employed in the framework are dictionary-based and machine learning. While the former approach is commonly employed the latter is not used for aggregating public sentiment. In this framework we explored how the machine learning approach can be used like the dictionary-based approach for analysing public sentiment. One case study, namely the Royal Birth of 2013 in the UK, was considered to compare the public sentiment scores estimated by the two approaches. Preliminary efforts indicate that there is a reasonable correlation between scores produced by the two approaches and indicate the feasibility of the machine learning approach for analysing public sentiment.

A key observation from the case study is that the problem of managing and visualising tweets for events that span across days cannot be maintained and analysed using traditional databases and data management techniques. For example, the tweet corpus for a two day period contained nearly one million tweets resulting in approximately five gigabytes of data. Such large amounts of data will require `big data' techniques, such as the use of Hadoop to address the data processing challenge. Faster methods will need to be developed to facilitate real-time analysis and visualisation of public sentiment. The machine learning approach is a slow method compared to the dictionary-based approach and in this research could not be employed for real-time visualisation as an event was unfolding. While the framework is capable of rapidly ingesting data, it cannot process data rapidly. Again fast and parallel methods for processing will need to be explored. 

Looking forward, this research aims to progress in the direction of employing big data techniques and parallel methods to develop a framework for real-time analysis and visualisation of public sentiment. Methods will be pursued to analyse tweets for capturing a broader spectrum of sentiments. Efforts will also be made towards developing a distributed framework available for public use.

\end{document}